\documentclass[conference]{IEEEtran}
\IEEEoverridecommandlockouts

\usepackage{tabularx}  
\usepackage{booktabs} 
\usepackage{graphicx} 
\usepackage{subcaption} 
\usepackage{hyperref}  
\hypersetup{  
    colorlinks=true, 
    linkcolor=red,   
    citecolor=green, 
    urlcolor=blue,   
    filecolor=magenta,
    linktoc=all      
} 

\usepackage{cite}
\usepackage{authblk}  
\usepackage{amsmath,amssymb,amsfonts}
\usepackage{algorithmic}
\usepackage{graphicx}
\usepackage{textcomp}
\usepackage{xcolor}
\def\BibTeX{{\rm B\kern-.05em{\sc i\kern-.025em b}\kern-.08em
    T\kern-.1667em\lower.7ex\hbox{E}\kern-.125emX}}
\begin{document}

\title{A Multi-annotated and Multi-modal Dataset for Wide-angle Video Quality Assessment\\
\thanks{$^{\dagger}$ Corresponding author. This work was supported in part by the Chongqing Natural Science Foundation Innovation and Development Joint Fund Project under Grant CSTB2023NSCQ-LZX0085, in part by the National Natural Science Foundation of China under Grants 62101084, 62471349 and 62171340, in part by the Science and Technology Research Program of Chongqing Municipal Education Commission under Grant KJQN202200638, in part by the Chongqing Postdoctoral Research Project Special Support under Grant 2022CQBSHTB2052, in part by the Natural Science Foundation of Chongqing under Grant CSTB2023NSCQ-BHX0187.}
}

\author[1]{Bo Hu}  
\author[1]{Wei Wang}  
\author[1]{Chunyi Li} 
\author[2]{Lihuo He}  
\author[3]{Leida Li}
\author[1]{Xinbo Gao$^{\dagger}$}  
  
\affil[1]{Key Laboratory of Image Cognition, Chongqing University of Posts and Telecommunications, Chongqing, China}  
\affil[2]{School of Electronic Engineering, Xidian University, Xi'an, China}
\affil[3]{School of Artificial Intelligence, Xidian University, Xi’an, China}
\maketitle


\begin{abstract}
Wide-angle video is favored for its wide viewing angle and ability to capture a large area of scenery, making it an ideal choice for sports and adventure recording. However, wide-angle video is prone to deformation, exposure and other distortions, resulting in poor video quality and affecting the perception and experience, which may seriously hinder its application in fields such as competitive sports. Up to now, few explorations focus on the quality assessment issue of wide-angle video. This deficiency primarily stems from the absence of a specialized dataset for wide-angle videos. To bridge this gap, we construct the first Multi-annotated and multi-modal Wide-angle Video quality assessment (MWV) dataset. Then, the performances of state-of-the-art video quality methods on the MWV dataset are investigated by inter-dataset testing and intra-dataset testing. Experimental results show that these methods impose significant limitations on their applicability.
\end{abstract}


\section{Introduction}
\begin{figure}[t]
    \centering
    \includegraphics[width=1\linewidth]{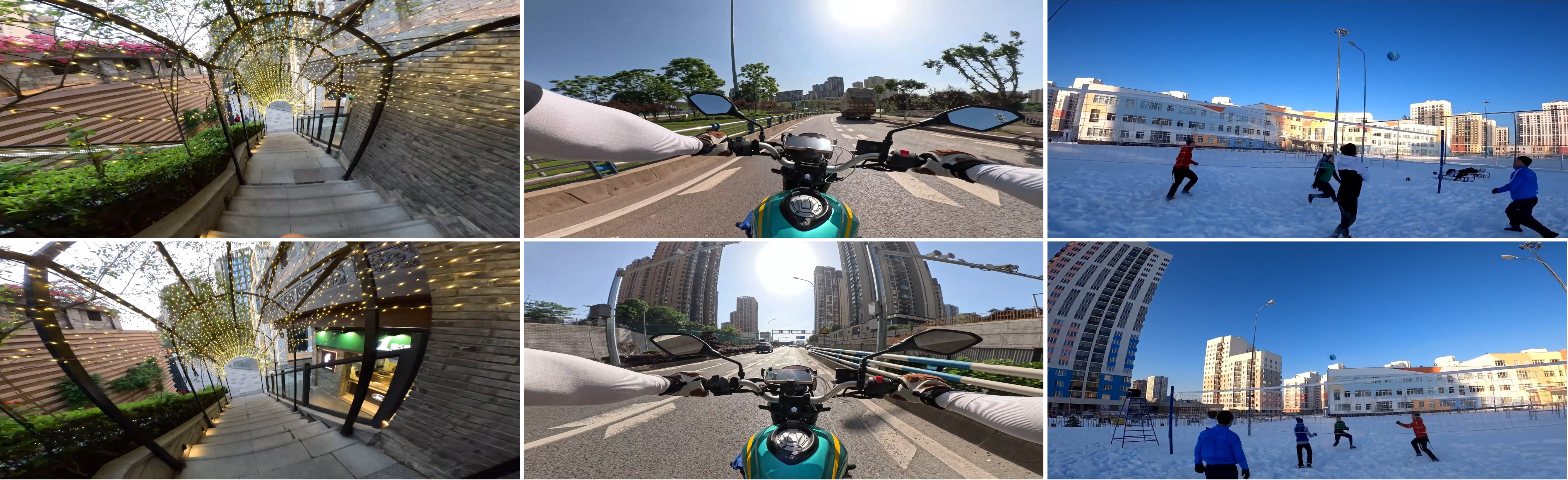}
    \caption{Example video clips in the MWV dataset.}
    \label{fig1}
\end{figure}

The market size of wide-angle cameras is projected to be \$4.41 billion in 2024 and is expected to reach \$9.18 billion by 2029, with a compound annual growth rate of 15.80\% during the forecast period (2024-2029) \cite{website:example}. Compared to digital single lens reflex and standard cameras, wide-angle cameras offer multiple advantages, such as easily recording high-resolution videos, capturing extreme sports like surfing and skiing with a wider field of view, and serving as dash cameras. Therefore, wide-angle cameras are anticipated to have a significantly larger user base in the future. For example, when users want to obtain a wider field of view to capture beautiful scenery while cycling outdoors, they often choose wide-angle or even ultra-wide-angle modes. On the other hand, wide-angle videos are often prone to  deformation, exposure and so on. As shown in Fig. \ref{fig1}, the images above and below are frame slices from the same video. It can be observed that the deformation effects are different. The set of images below exhibits more severe deformation, which will seriously affect people's perception and experience. Therefore, it is necessary to deeply study the quality assessment of wide-angle video.

\newcolumntype{Y}{>{\centering\arraybackslash}p{0.158\textwidth}} 
\begin{table*}[t]    
\centering    
\caption{Summary of popular public video quality datasets.}    
\begin{tabularx}{\textwidth}{@{}Y|Y|c|Y|c|c|c|c@{}}    
\toprule    
Dataset & Modal & Deformation & Video Resolution & Frame Rate & Video Contents & Duration & Format \\     
\midrule    
CVD2014\cite{nuutinen2016cvd2014} & Video & No & 480P \& 720P & 9-30 & 234 & 10-25 & AVI \\        
KoNViD-1K\cite{hosu2017konstanz} & Video & No & 540P & 24-30 & 1200 & 8 & MP4 \\        
LBVD\cite{chen2019qoe} & Video & No & 240P-540P & - & 1013 & 10 & MP4 \\        
LIVE-VQC\cite{sinno2018large} & Video & No & 240P-1080P & 19-30 & 585 & 10 & MP4 \\        
LSVQ\cite{ying2021patch} & Video & No & 99P-4K & - & 39075 & 5-12 & MP4 \\     
\midrule    
MWV & Video \& Text & Yes & 720P-4K & 30 \& 60 & 1000 & 10 & MP4 \\     
\bottomrule    
\end{tabularx}    
\label{table 1}    
\end{table*}  

At present, due to the extensive attention and efforts of the community, the subjective and objective assessment of video quality has made great progress. For subjective assessment, a large number of datasets are proposed, including LSVQ \cite{ying2021patch}, KoNViD-1K \cite{hosu2017konstanz}, CVD2014 \cite{nuutinen2016cvd2014}, LIVE-VQC \cite{sinno2018large} and LBVD \cite{chen2019qoe}. Despite the fact that the aforementioned datasets incorporate authentic distortions, they do not specifically address video distortions related to deformations. The general characteristics of these video datasets are listed in Table \ref{table 1}. In addition, few studies have focused on the subjective quality assessment of wide-angle videos. So far, our best efforts have not found a publicly available dataset. For objective assessment, although some no-reference quality assessment methods perform well on the above datasets, their performance for wide-angle video is unknown. Therefore, it is necessary to construct a new dataset of wide-angle video quality assessment to promote the rapid development of this field.

Video quality perception is a complex subjective process that involves assessing and interpreting visual stimuli. When perceiving visual quality, humans can simultaneously obtain a variety of perceptual information \cite{baltruvsaitis2018multimodal} \cite{pinson2012influence}. Our brains can easily establish connections between different modalities during perception, thus gaining a more comprehensive understanding of things \cite{song2019harmonized}. For example, when we watch a skiing video, our brain will involuntarily generate various linguistic descriptions, providing various descriptive clues for quality perception. Inspired by this, we construct a new Multi-annotated and multi-modal Wide-angle Video quality assessment (MWV) dataset. To the best of our knowledge, the proposed MWV dataset is the first wide-angle video quality assessment dataset. Firstly, wide-angle videos are collected through both downloading and filming. Secondly, a subjective experiment is conducted using the single stimulus continuous quality evaluation method. Then, the experimental data is subjected to outlier removal and analysis. Finally, intra-database experiments and cross-database experiments are performed on the MWV dataset. This effort aims to drive the development of objective quality assessment methods to meet the demand for high-quality wide-angle video experiences among users today. 

\section{Establishment of The Dataset}
In this section, the MWV dataset is constructed through steps video collection, subjective experiment, and data processing, etc. Finally, we prove the rationality of dataset construction through data analysis.
 
\subsection{Wide-angle Videos Collection}
\begin{figure}[t]
    \centering
    \includegraphics[width=1\linewidth]{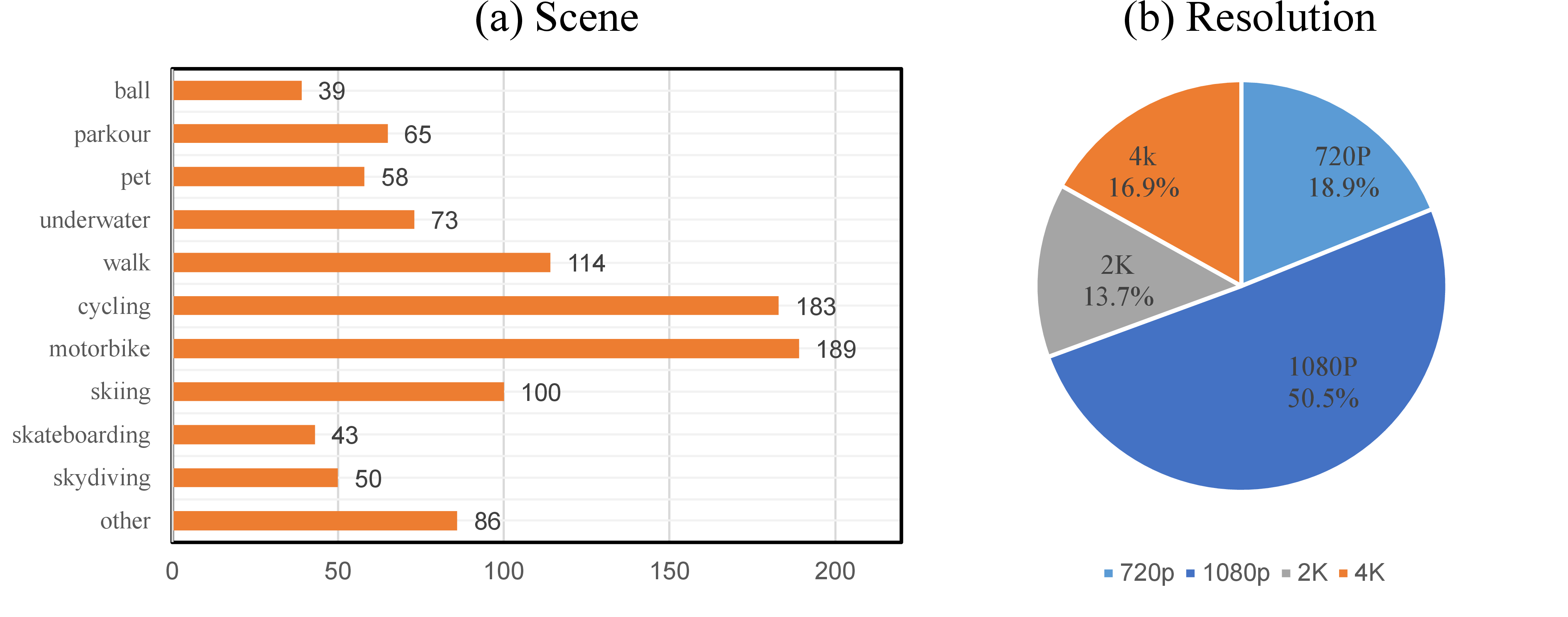}
    \caption{A statistical analysis of the MWV dataset. (a) video scene, (b) video resolution.} 
    \label{fig2}
\end{figure}

To collect wide-angle videos from various scenes, we refer to the recommended usage scenarios published on the official websites of companies such as GoPro, DJI, and Insta360. We then search for these recommended scenarios on social media platforms like Bilibili and TikTok, recording the number of search results to ensure the subsequent data distribution aligns with current internet trends. To enhance the dataset's challenge, we, as amateur videographers, utilize a range of devices for filming. Following the suggestions in \cite{frohlich2012qoe}, we manually select 1,000 challenging wide-angle videos and trim them into 10-second clips while considering content continuity. Crop the original video without changing its resolution, but only slice it in time. Fig. \ref{fig2} illustrates the data distribution of scene categories and resolutions within the dataset. It can be seen from Fig. \ref{fig2} that these videos are diverse in content and resolution, laying the foundation for successfully building MWV dataset.

\subsection{Subjective Experiment}
\begin{figure}[t]
    \centering
    \includegraphics[width=1\linewidth]{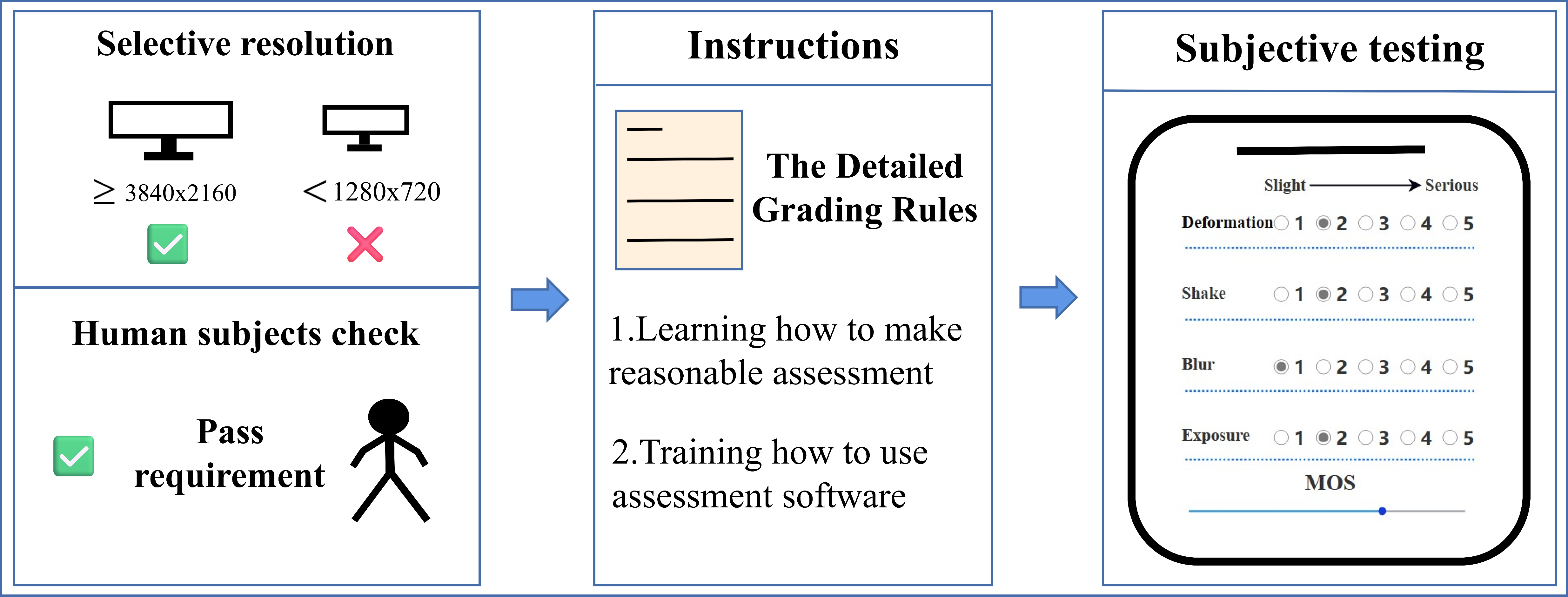}
    \caption{Subjective study workflow.}
    \label{fig3}
\end{figure}

The human intelligence task (HIT) pipeline is shown in Fig. \ref{fig3}. Before each task begins, equipment checks are performed, and participants' physical and psychological states are assessed. Participants are required to read a general set of instructions, followed by a related test to verify their understanding of these instructions. They must pass this test to proceed. Finally, a total of 22 participants, half male and half female, took part in our study.

\begin{figure}[t]
    \centering
    \includegraphics[width=1\linewidth]{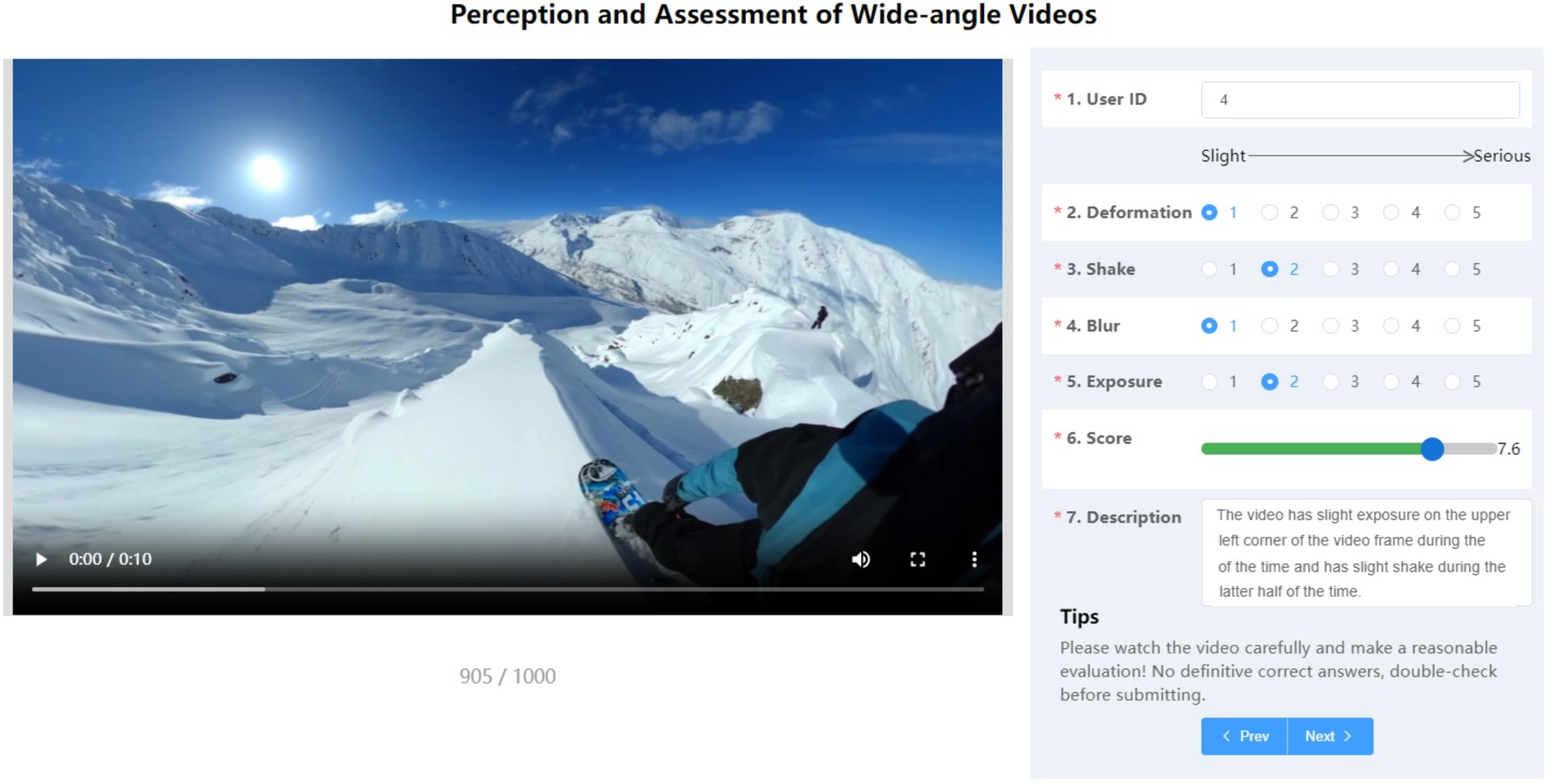}
    \caption{Graphical interface of subjective experiment.}
    \label{fig4}
\end{figure}

According to ITU-R BT.500-14 \cite{bt2019500}, ITU-T Rec. P.910 \cite{shang2023study} and NNID \cite{xiang2019blind}, we employ the single stimulus method and design a graphical user interface (Fig. \ref{fig4}) to conduct the experiments. Before commencing the experiment, it is ensure that all videos are downloaded locally to avoid any disruptions caused by network connectivity issues. The graphical interface scale is adjusted to 100\% to prevent any scaling artifacts during video playback. To avoid participant fatigue, the entire experiment is divided into 20 separate viewing sessions, each containing 50 videos, with a 24-hour interval between each session \cite{cheng2020screen}. 

Participants are required to rate four attributes: deformation, shake, blur, and exposure. Each attribute is rated on a scale from 1 to 5, representing the severity of the attribute from minor to severe. Specifically, a rating of 1 indicates without distortion, 2 indicates slight distortion, 3 indicates moderate distortion, 4 indicates serious distortion with very noticeable distortion, and 5 indicates extreme distortion with highly severe distortion. Participants provide a quality score corresponding to their perceived quality by dragging a slider (ranging from 0 to 10). 

\begin{figure}[h]
    \centering
    \includegraphics[width=1\linewidth]{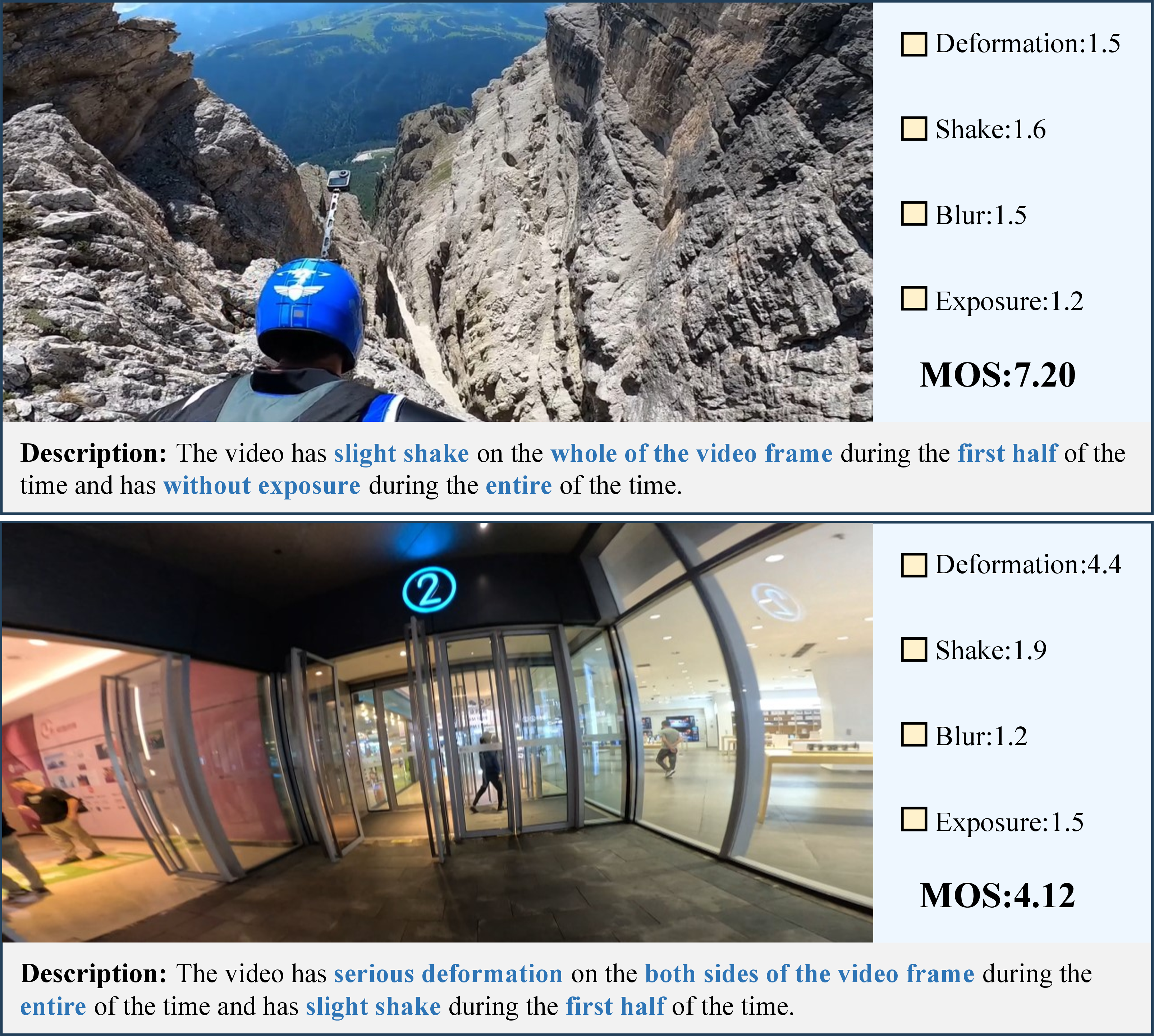}
    \caption{An example of the multi-annotated results and the textual description.}
    \label{fig6}
\end{figure}

Linguistic description helps to avoid personal information bias, because the possibility of inconsistent linguistic description is far less than that of consistency \cite{wade2013visual} \cite{hanjalic2005affective} \cite{wang2023blind}. In order to make up for the limitations of most existing video quality assessment methods (which tend to focus only on a single mode) \cite{zhang2023md} \cite{chen2020rirnet} \cite{venkataramanan2023one} \cite{shen2023blind}, we further describe the video perceptual quality in four aspects: deformation, shake, blur, and exposure. To address the above four challenges, we attempt to develop a tractable verbally description paradigm. A quality-based sentence should contain at least two set of perceptual attributes, and provide a meaningful and concise description of the degree of distortion, distortion location, and distortion time period for each attribute. Fig. \ref{fig6} shows an example of the multi-annotated results and the textual description.

\subsection{Processing of Raw Data}
We regularly remove 2 subjects based on the outlier removal method outlined in \cite{sheikh2006statistical} \cite{hu2016perceptual} after conducting subjective experiments. Subsequently, we discard all subjective data reported by these rejected subjects, resulting in a final count of 20 valid participants.

First, we calculate the average subjective score for each video:
\begin{equation}
    \overline{s }=\frac{1}{N}\displaystyle\sum_{i=1}^{N}s_i
\end{equation}
where $s_i$ is the subjective score of the $i$th subject, $N$ is the number of subjects. Then,  the standard deviation is calculated \cite{bt2019500}:
\begin{equation}
    \sigma=\sqrt{\displaystyle\sum_{i=1}^{N}\frac{(s_i-\overline{s})^2}{N-1}}
\end{equation}

Assuming that the subjective scores follow a normal distribution, we then calculate the confidence interval to identify outliers, namely $[\overline{s }-\delta ,\overline{s }+\delta]$. For this work, the confidence level is set at 95\%. Consequently, according to \cite{bt2019500}, we derive $\delta =1.96*\frac{\sigma }{\sqrt{N}}$. For the confidence interval, if a score falls outside the range of  $[\overline{s }-\delta ,\overline{s }+\delta]$, it is considered an outlier and is subsequently removed. Finally, the average of the remaining subjective scores is calculated as the Mean Opinion Score (MOS).

\subsection{Data Analysis}

\begin{figure}[h]
    \centering
    \includegraphics[width=1\linewidth]{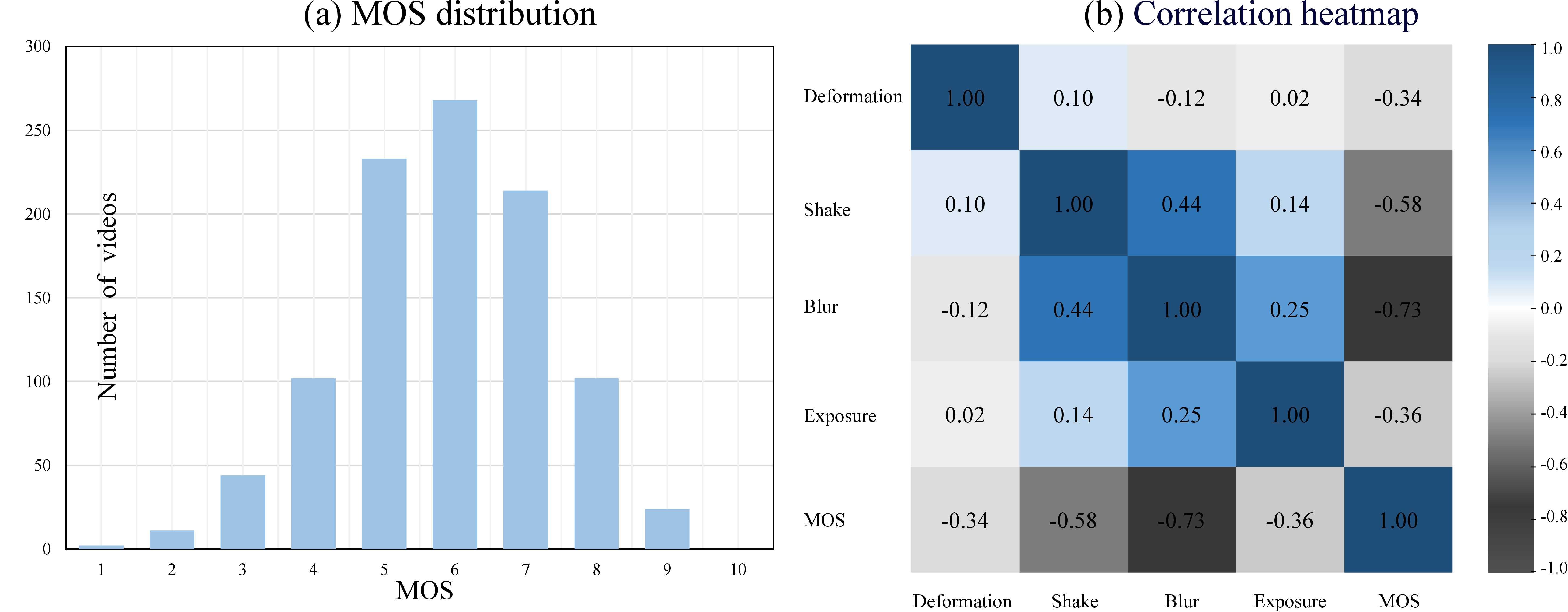}
    \caption{A statistical analysis of the MWV dataset. (a) MOS distribution, (b) heatmap of correlation between distortion type and MOS.} 
    \label{fig5}
\end{figure}

Fig. \ref{fig5} shows the histogram distribution of the final MOS values of 1000 videos across 10 equally spaced bins, along with the correlation between four distortion types and MOS values. As depicted in Fig. \ref{fig5}(a), the MOS distribution exhibits a Gaussian distribution within the range of [0,10], indicating the rationality of the proposed MWV dataset. As shown in Fig. \ref{fig5}(b), the correlation heatmap visually demonstrates the negative correlation between the four types of distortion and MOS values. There is no significant correlation between shake, exposure, and blur. However, there is a certain degree of correlation between shake and blur, which is easily understood as shake can cause corresponding motion blur. The above data analysis further proves the rationality and correctness of the proposed MWV dataset.

\section{Experiments}
\label{sec:pagestyle}
\subsection{Implementation Details}
Pearson Linear Correlation Coefficient (PLCC) \cite{lee1988thirteen} and Spearman Rank Correlation Coefficient (SRCC) \cite{kendall1948rank} are used as the experimental metrics. All experiments are conducted on two NVIDIA GeForce RTX 4090 GPUs. Experiments on each database are processed through k-fold (k = 10) cross-validation, ensuring that the training sets and test sets do not overlap in content \cite{wen2024modular}, and the average values are taken as the experimental results \cite{chen2021unsupervised}. The hyperparameters of all methods are maintained identical to those reported in the original papers to ensure the algorithms' performance advantages are preserved.

\begin{table}[t]    
\centering    
\caption{Inter-dataset performance of existing representative VQA methods.}    
\renewcommand{\arraystretch}{1.3} 
\begin{tabular}{l|c|c|c|c}    
\hline    
\textbf{Training dataset:} & \multicolumn{2}{c|}{\textbf{LSVQ}} & \multicolumn{2}{c}{\textbf{LSVQ}} \\ \hline    
\textbf{Testing dataset:}  & \multicolumn{2}{c|}{\textbf{LSVQ}} & \multicolumn{2}{c}{\textbf{MWV}} \\ \hline    
Method            & \multicolumn{1}{c|}{SRCC} &PLCC & \multicolumn{1}{c|}{SRCC} & PLCC \\ \hline    
VGG-19     (ICLR, 2015)\cite{simonyan2014very}  & 0.816 & 0.815                      & 0.568         & 0.640         \\    
Resnet-50  (CVPR, 2016)\cite{he2016deep}        & 0.826 & 0.827                      & 0.594         & 0.646         \\    
SimpleVQA (ACMMM, 2022)\cite{sun2022deep}       & 0.867 & 0.861                      & 0.601         & 0.657         \\    
DSD-PRO   (TCSVT, 2023)\cite{liu2022quality}    & 0.869 & 0.868                      & 0.603         & 0.665         \\    
FastVQA    (ECCV, 2022)\cite{wu2022fast}        & 0.876 & 0.877                      & 0.571         & 0.621         \\    
DOVER      (ICCV, 2023)\cite{wu2023exploring}   & 0.888 & 0.889                      & 0.611         & 0.668         \\ \hline    
\end{tabular}    
\label{table 2}  
\end{table}

\subsection{Inter-dataset Performance}
Whether the existing Video Quality Assessment (VQA) methods can be directly applied to wide-angle video has not been studied, which concerns the necessity of constructing wide-angle VQA dataset. To this end, this subsection uses LSVQ and MWV datasets to explore the performance of existing quality methods to directly predict wide-angle video quality. Specifically, four representative VQA methods are selected to test the performance, they are SimpleVQA \cite{sun2022deep}, DSD-PRO \cite{liu2022quality}, FastVQA \cite{wu2022fast}, and DOVER \cite{wu2023exploring}. These methods are trained on the largest existing video quality dataset, LSVQ, and tested on the proposed MWV dataset. The results are shown in Table \ref{table 2}. 

As can be seen from the table, the existing VQA methods already perform well on the LSVQ dataset. However, even with the support of the largest dataset in the current video quality assessment field, these VQA methods do not perform well on the wide-angle video dataset. The reason is that compared with standard video, wide-angle video is particularly prone to severe deformation and violent shake, which significantly affect the user experience and are not considered in the current algorithm design. Therefore, to evaluate the quality of wide-angle videos, it is essential to first construct a wide-angle video quality assessment dataset.

\subsection{Intra-dataset Performance}
The intra-dataset performance is evaluated on the MWV dataset, with 80\% for training and 20\% for testing \cite{zhu2022learning} \cite{liu2021spatiotemporal} \cite{chen2021learning} \cite{cao2023subjective}. The results are shown in Table \ref{table 3}. The table reveals that these quality methods perform better when trained and tested on the MWV dataset compared to being trained on LSVQ and tested on the MWV dataset. However, even when trained and tested on the MWV dataset, the performance of these existing VQA methods do not achieve very good results, indicating that they are unable to extract the unique features of wide-angle videos. This highlights the need to develop specialized quality methods for wide-angle video quality assessment. Therefore, constructing a wide-angle video dataset is crucial for advancing the development of methods specifically aimed at evaluating the quality of wide-angle videos.

\begin{table}[t]
\caption{Intra-dataset performance of existing representative VQA methods.} 
\renewcommand{\arraystretch}{1.3} 
\scalebox{1}{ 
\begin{tabular}{l|cc|cc|cc}
\hline
\textbf{Dataset:} & \multicolumn{2}{c|}{\textbf{MWV}} & \multicolumn{2}{c|}{\textbf{MWV-slight}} & \multicolumn{2}{c}{\textbf{MWV-serious}} \\ \hline
Method            & \multicolumn{1}{c|}{SRCC}    & PLCC    & \multicolumn{1}{c|}{SRCC}     & PLCC    & \multicolumn{1}{c|}{SRCC}     & PLCC    \\ \hline
VGG19              & \multicolumn{1}{c|}{0.718}   & 0.762   & \multicolumn{1}{c|}{0.721}    & 0.755   & \multicolumn{1}{c|}{0.494}    & 0.683   \\
Resnet50           & \multicolumn{1}{c|}{0.745}   & 0.783   & \multicolumn{1}{c|}{0.757}    & 0.812   & \multicolumn{1}{c|}{0.455}    & 0.584   \\
SimpleVQA          & \multicolumn{1}{c|}{0.747}   & 0.785   & \multicolumn{1}{c|}{0.765}    & 0.825   & \multicolumn{1}{c|}{0.562}    & 0.671   \\
DSD-PRO            & \multicolumn{1}{c|}{0.773}   & 0.801   & \multicolumn{1}{c|}{0.784}    & 0.808   & \multicolumn{1}{c|}{0.503}    & 0.595   \\
FastVQA            & \multicolumn{1}{c|}{0.774}   & 0.819   & \multicolumn{1}{c|}{0.805}    & 0.837   & \multicolumn{1}{c|}{0.596}    & 0.697   \\
DOVER              & \multicolumn{1}{c|}{0.782}   & 0.824   & \multicolumn{1}{c|}{0.816}    & 0.843   & \multicolumn{1}{c|}{0.607}    & 0.696   \\ \hline
\end{tabular}
}
\label{table 3}
\end{table}

To verify the performance of these quality methods under different degrees of deformation, we further divide the dataset into two subsets. Specifically, we defined the interval [1,3) as the slightly deformed group (MWV-slight), and the interval [3,5] as the seriously deformed group (MWV-serious). We then trained and tested these methods on these two groups of videos, respectively. The results are shown in Table \ref{table 3}. From the table, it can be seen that the SRCC and PLCC values are highest for the slightly deformed group, followed by the entire wide-angle video group, and lowest for the seriously deformed group. This further illustrates the limitations of existing  quality  methods for wide-angle videos. While these methods can handle slightly deformed videos relatively effectively, they do not perform well in heavily distorted videos. This further emphasizes the necessity and urgency of constructing a wide-angle video dataset to lay the foundation for the development of specialized wide-angle video quality assessment methods.

\section{Conclusion}
In this paper, the first multi-annotated and multi-modal wide-angle video quality dataset has been constructed, aiming to explore the characteristics of wide-angle videos in the field of video quality assessment. The differences between wide-angle videos and traditional standard videos in quality assessment has been validated from both subjective and objective perspectives. Extensive subjective experiments has been conducted  to comprehensively evaluate wide-angle videos in terms of attribute annotations, quality scores, and textual descriptions. The results of these subjective experiments has been analyzed to prove the correctness and rationality of the dataset construction. For objective assessment, existing state-of-the-art methods has been used to conduct experimental analysis on wide-angle videos in terms of inter-dataset performance and intra-dataset performance. The need for further development of quality methods specifically tailored for wide-angle videos has been indicated by the objective analysis.

\bibliographystyle{IEEEbib}
\bibliography{refs}

\begin{thebibliography}{10}

\bibitem{website:example}
``Mordorintelligence,'' [Online] Available,
\newblock {https://www.mordorintelligence.com/zh-CN/industry-reports/action-camera-market}.

\bibitem{nuutinen2016cvd2014}
Mikko Nuutinen, Toni Virtanen, Mikko Vaahteranoksa, Tero Vuori, Pirkko Oittinen, and Jukka H{\"a}kkinen,
\newblock ``Cvd2014—a database for evaluating no-reference video quality assessment algorithms,''
\newblock {\em IEEE Transactions on Image Processing}, vol. 25, no. 7, pp. 3073--3086, 2016.

\bibitem{hosu2017konstanz}
Vlad Hosu, Franz Hahn, Mohsen Jenadeleh, Hanhe Lin, Hui Men, Tam{\'a}s Szir{\'a}nyi, Shujun Li, and Dietmar Saupe,
\newblock ``The konstanz natural video database (konvid-1k),''
\newblock in {\em 2017 Ninth International Conference on Quality of Multimedia Experience (QoMEX)}. IEEE, 2017, pp. 1--6.

\bibitem{chen2019qoe}
Pengfei Chen, Leida Li, Yipo Huang, Fengfeng Tan, and Wenjun Chen,
\newblock ``Qoe evaluation for live broadcasting video,''
\newblock in {\em 2019 IEEE International Conference on Image Processing (ICIP)}. IEEE, 2019, pp. 454--458.

\bibitem{sinno2018large}
Zeina Sinno and Alan~Conrad Bovik,
\newblock ``Large-scale study of perceptual video quality,''
\newblock {\em IEEE Transactions on Image Processing}, vol. 28, no. 2, pp. 612--627, 2018.

\bibitem{ying2021patch}
Zhenqiang Ying, Maniratnam Mandal, Deepti Ghadiyaram, and Alan Bovik,
\newblock ``Patch-vq:'patching up'the video quality problem,''
\newblock in {\em Proceedings of the IEEE/CVF Conference on Computer Vision and Pattern Recognition}, 2021, pp. 14019--14029.

\bibitem{baltruvsaitis2018multimodal}
Tadas Baltru{\v{s}}aitis, Chaitanya Ahuja, and Louis-Philippe Morency,
\newblock ``Multimodal machine learning: A survey and taxonomy,''
\newblock {\em IEEE Transactions on Pattern Analysis and Machine Intelligence}, vol. 41, no. 2, pp. 423--443, 2018.

\bibitem{pinson2012influence}
Margaret~H Pinson, Lucjan Janowski, Romuald P{\'e}pion, Quan Huynh-Thu, Christian Schmidmer, Phillip Corriveau, Audrey Younkin, Patrick Le~Callet, Marcus Barkowsky, and William Ingram,
\newblock ``The influence of subjects and environment on audiovisual subjective tests: An international study,''
\newblock {\em IEEE Journal of Selected Topics in Signal Processing}, vol. 6, no. 6, pp. 640--651, 2012.

\bibitem{song2019harmonized}
Guoli Song, Shuhui Wang, Qingming Huang, and Qi~Tian,
\newblock ``Harmonized multimodal learning with gaussian process latent variable models,''
\newblock {\em IEEE Transactions on Pattern Analysis and Machine Intelligence}, vol. 43, no. 3, pp. 858--872, 2019.

\bibitem{frohlich2012qoe}
Peter Fr{\"o}hlich, Sebastian Egger, Raimund Schatz, Michael M{\"u}hlegger, Kathrin Masuch, and Bruno Gardlo,
\newblock ``Qoe in 10 seconds: Are short video clip lengths sufficient for quality of experience assessment?,''
\newblock in {\em 2012 Fourth International Workshop on Quality of Multimedia Experience}. IEEE, 2012, pp. 242--247.

\bibitem{bt2019500}
ITU~Recommendation BT,
\newblock ``500-14, methodologies for the subjective assessment of the quality of television images,''
\newblock {\em Geneva: International Telecommunication Union}, 2019.

\bibitem{shang2023study}
Zaixi Shang, Joshua~P Ebenezer, Abhinau~K Venkataramanan, Yongjun Wu, Hai Wei, Sriram Sethuraman, and Alan~C Bovik,
\newblock ``A study of subjective and objective quality assessment of hdr videos,''
\newblock {\em IEEE Transactions on Image Processing}, vol. 33, pp. 42--57, 2023.

\bibitem{xiang2019blind}
Tao Xiang, Ying Yang, and Shangwei Guo,
\newblock ``Blind night-time image quality assessment: Subjective and objective approaches,''
\newblock {\em IEEE Transactions on Multimedia}, vol. 22, no. 5, pp. 1259--1272, 2019.

\bibitem{cheng2020screen}
Shan Cheng, Huanqiang Zeng, Jing Chen, Junhui Hou, Jianqing Zhu, and Kai-Kuang Ma,
\newblock ``Screen content video quality assessment: Subjective and objective study,''
\newblock {\em IEEE Transactions on Image Processing}, vol. 29, pp. 8636--8651, 2020.

\bibitem{wade2013visual}
Nicholas Wade and Mike Swanston,
\newblock {\em Visual perception: An introduction},
\newblock Psychology Press, 2013.

\bibitem{hanjalic2005affective}
Alan Hanjalic and Li-Qun Xu,
\newblock ``Affective video content representation and modeling,''
\newblock {\em IEEE Transactions on Multimedia}, vol. 7, no. 1, pp. 143--154, 2005.

\bibitem{wang2023blind}
Miaohui Wang, Zhuowei Xu, Mai Xu, and Weisi Lin,
\newblock ``Blind multimodal quality assessment of low-light images,''
\newblock {\em arXiv preprint arXiv:2303.10369}, 2023.

\bibitem{zhang2023md}
Zicheng Zhang, Wei Wu, Wei Sun, Danyang Tu, Wei Lu, Xiongkuo Min, Ying Chen, and Guangtao Zhai,
\newblock ``Md-vqa: Multi-dimensional quality assessment for ugc live videos,''
\newblock in {\em Proceedings of the IEEE/CVF Conference on Computer Vision and Pattern Recognition}, 2023, pp. 1746--1755.

\bibitem{chen2020rirnet}
Pengfei Chen, Leida Li, Lei Ma, Jinjian Wu, and Guangming Shi,
\newblock ``Rirnet: Recurrent-in-recurrent network for video quality assessment,''
\newblock in {\em Proceedings of the 28th ACM International Conference on Multimedia}, 2020, pp. 834--842.

\bibitem{venkataramanan2023one}
Abhinau~K Venkataramanan, Cosmin Stejerean, Ioannis Katsavounidis, and Alan~C Bovik,
\newblock ``One transform to compute them all: Efficient fusion-based full-reference video quality assessment,''
\newblock {\em IEEE Transactions on Image Processing}, 2023.

\bibitem{shen2023blind}
Wenhao Shen, Mingliang Zhou, Xuekai Wei, Heqiang Wang, Bin Fang, Cheng Ji, Xu~Zhuang, Jason Wang, Jun Luo, Huayan Pu, et~al.,
\newblock ``A blind video quality assessment method via spatiotemporal pyramid attention,''
\newblock {\em IEEE Transactions on Broadcasting}, 2023.

\bibitem{sheikh2006statistical}
Hamid~R Sheikh, Muhammad~F Sabir, and Alan~C Bovik,
\newblock ``A statistical evaluation of recent full reference image quality assessment algorithms,''
\newblock {\em IEEE Transactions on Image Processing}, vol. 15, no. 11, pp. 3440--3451, 2006.

\bibitem{hu2016perceptual}
Bo~Hu, Leida Li, Jiansheng Qian, and Yuming Fang,
\newblock ``Perceptual evaluation of compressive sensing image recovery,''
\newblock in {\em 2016 Eighth International Conference on Quality of Multimedia Experience (QoMEX)}. IEEE, 2016, pp. 1--6.

\bibitem{lee1988thirteen}
Joseph Lee~Rodgers and W~Alan Nicewander,
\newblock ``Thirteen ways to look at the correlation coefficient,''
\newblock {\em The American Statistician}, vol. 42, no. 1, pp. 59--66, 1988.

\bibitem{kendall1948rank}
Maurice~George Kendall,
\newblock ``Rank correlation methods.,''
\newblock 1948.

\bibitem{wen2024modular}
Wen Wen, Mu~Li, Yabin Zhang, Yiting Liao, Junlin Li, Li~Zhang, and Kede Ma,
\newblock ``Modular blind video quality assessment,''
\newblock in {\em Proceedings of the IEEE/CVF Conference on Computer Vision and Pattern Recognition}, 2024, pp. 2763--2772.

\bibitem{chen2021unsupervised}
Pengfei Chen, Leida Li, Jinjian Wu, Weisheng Dong, and Guangming Shi,
\newblock ``Unsupervised curriculum domain adaptation for no-reference video quality assessment,''
\newblock in {\em Proceedings of the IEEE/CVF International Conference on Computer Vision}, 2021, pp. 5178--5187.

\bibitem{simonyan2014very}
Karen Simonyan and Andrew Zisserman,
\newblock ``Very deep convolutional networks for large-scale image recognition,''
\newblock {\em arXiv preprint arXiv:1409.1556}, 2014.

\bibitem{he2016deep}
Kaiming He, Xiangyu Zhang, Shaoqing Ren, and Jian Sun,
\newblock ``Deep residual learning for image recognition,''
\newblock in {\em Proceedings of the IEEE Conference on Computer Vision and Pattern Recognition}, 2016, pp. 770--778.

\bibitem{sun2022deep}
Wei Sun, Xiongkuo Min, Wei Lu, and Guangtao Zhai,
\newblock ``A deep learning based no-reference quality assessment model for ugc videos,''
\newblock in {\em Proceedings of the 30th ACM International Conference on Multimedia}, 2022, pp. 856--865.

\bibitem{liu2022quality}
Yongxu Liu, Jinjian Wu, Leida Li, Weisheng Dong, and Guangming Shi,
\newblock ``Quality assessment of ugc videos based on decomposition and recomposition,''
\newblock {\em IEEE Transactions on Circuits and Systems for Video Technology}, vol. 33, no. 3, pp. 1043--1054, 2022.

\bibitem{wu2022fast}
Haoning Wu, Chaofeng Chen, Jingwen Hou, Liang Liao, Annan Wang, Wenxiu Sun, Qiong Yan, and Weisi Lin,
\newblock ``Fast-vqa: Efficient end-to-end video quality assessment with fragment sampling,''
\newblock in {\em European Conference on Computer Vision}. Springer, 2022, pp. 538--554.

\bibitem{wu2023exploring}
Haoning Wu, Erli Zhang, Liang Liao, Chaofeng Chen, Jingwen Hou, Annan Wang, Wenxiu Sun, Qiong Yan, and Weisi Lin,
\newblock ``Exploring video quality assessment on user generated contents from aesthetic and technical perspectives,''
\newblock in {\em Proceedings of the IEEE/CVF International Conference on Computer Vision}, 2023, pp. 20144--20154.

\bibitem{zhu2022learning}
Hanwei Zhu, Baoliang Chen, Lingyu Zhu, and Shiqi Wang,
\newblock ``Learning spatiotemporal interactions for user-generated video quality assessment,''
\newblock {\em IEEE Transactions on Circuits and Systems for Video Technology}, vol. 33, no. 3, pp. 1031--1042, 2022.

\bibitem{liu2021spatiotemporal}
Yongxu Liu, Jinjian Wu, Leida Li, Weisheng Dong, Jinpeng Zhang, and Guangming Shi,
\newblock ``Spatiotemporal representation learning for blind video quality assessment,''
\newblock {\em IEEE Transactions on Circuits and Systems for Video Technology}, vol. 32, no. 6, pp. 3500--3513, 2021.

\bibitem{chen2021learning}
Baoliang Chen, Lingyu Zhu, Guo Li, Fangbo Lu, Hongfei Fan, and Shiqi Wang,
\newblock ``Learning generalized spatial-temporal deep feature representation for no-reference video quality assessment,''
\newblock {\em IEEE Transactions on Circuits and Systems for Video Technology}, vol. 32, no. 4, pp. 1903--1916, 2021.

\bibitem{cao2023subjective}
Yuqin Cao, Xiongkuo Min, Wei Sun, and Guangtao Zhai,
\newblock ``Subjective and objective audio-visual quality assessment for user generated content,''
\newblock {\em IEEE Transactions on Image Processing}, 2023.

\end{thebibliography}

\end{document}